\documentclass{article}

\usepackage{microtype}
\usepackage{graphicx}
\usepackage{subfigure}
\usepackage{amsmath}
\usepackage{booktabs} 

\usepackage{hyperref}

\usepackage{icml2019}

\icmltitlerunning{Spatio-temporal crop type classification with capsule layers and distributed attention}

\begin{document}

\twocolumn[
\icmltitle{Spatio-temporal crop classification of low-resolution satellite imagery with capsule layers and distributed attention}



\icmlsetsymbol{equal}{*}

\begin{icmlauthorlist}
\icmlauthor{John Brandt}{ya}
\end{icmlauthorlist}

\icmlaffiliation{ya}{Yale University}
\icmlcorrespondingauthor{Cieua Vvvvv}{c.vvvvv@googol.com}
\icmlcorrespondingauthor{Eee Pppp}{ep@eden.co.uk}

\icmlkeywords{Machine Learning, ICML}

\vskip 0.3in
]



\printAffiliationsAndNotice{\icmlEqualContribution} 

\begin{abstract}
Land use classification of low resolution spatial imagery is one of the most extensively researched fields in remote sensing. Despite significant advancements in satellite technology, high resolution imagery lacks global coverage and can be prohibitively expensive to procure for extended time periods. Accurately classifying land use change without high resolution imagery offers the potential to monitor vital aspects of global development agenda including climate smart agriculture, drought resistant crops, and sustainable land management. Utilizing a combination of capsule layers and long-short term memory layers with distributed attention, the present paper achieves state-of-the-art accuracy on temporal crop type classification at a 30x30m resolution with Sentinel 2 imagery. 
\end{abstract}

\section{Introduction}
\label{Introduction}

Over recent years, deep learning approaches to remote sensing have established new state of the art results for land cover classification. Pre-trained convolutional neural networks (CNNs) such as GoogleNet and CaffeNet have achieved remarkable performance on land use classification tasks when fine tuned with remote sensing data  \cite{Castelluccio2015LandUC}. In order to exploit the temporal nature of agriculture, Rubwurm and Korner \yrcite{RuBwurm2017TemporalVM} applied recurrent neural networks (RNN) and long short term memory networks (LSTM) to the task of temporal vegetation classification, finding that time series information greatly improved classification accuracy on multi-class vegetation data. 

Recently, capsule networks have been proposed for remote sensing classification tasks, though their applications primarily involve high-resolution satellite imagery. Deng et al. \yrcite{Deng2018HyperspectralIC} achieved a 20\% relative improvement in land use classification accuracy with capsule networks over CNN, additionally finding that capsule networks required far fewer labelled training data points to achieve high confidence in predictions. Paoletti et al. \yrcite{Paoletti2019CapsuleNF} also employ capsule networks to classify land use in high resolution satellite imagery, finding that the architecture greatly outperforms deep that of a deep CNN. 

In contrast to capsule networks, attention networks are relatively underused in the remote sensing field. Liu et al. \yrcite{Liu2018AttentionBN} developed a stacked LSTM network which reweighted later layers with attention over the input and the previous hidden state. Xu et al. \yrcite{xu_tao_lu_zhong_2018} incorporate attention layers over convolutional filters to create an embedding that combines weighted feature maps. However, these applications are primarily limited to high resolution spatial imagery and their applications to low resolution imagery are not well understood. This paper presents a new deep learning architecture (CapsAttn) which combines capsule and attention layers with CNN and LSTM layers to achieve state of the art results on temporal crop type classification with low-resolution satellite imagery.

\section{Methodology}

The data set used for this paper was provided in Rubworm and Korner \yrcite{RuBwurm2017TemporalVM}, comprising 26 Sentinel 2A images taken during 2016 of a 102 x 42 km area in Munich, Germany.  The 10 m bands (2 blue, 3 green, 4 red, 8 near-infrared) and 20 m bands (11 short-wave-infrared-1, 12 short-wave-infrared-2) down-sampled to 10 m resolution were extracted from 406,000 points of interest were within the data set. Input data was formatted to 3 x 3 px neighborhoods for each of the 26 time steps with an approximate 75, 5, and 20 percent train, validation, and test ratio. 

Ground truth labels were provided by the Bavarian Ministry of Agriculture, totalling 19 classes with at least 400 occurrences (corn, corn, meadow, asparagus, rape, hops, summer oats, winter spelt, fallow, winter wheat, winter barley, winter rye, beans, winter triticale, summer barley, peas, potatoes, soybeans, and sugar beets). Sample points were additionally classified into cloud, water, snow, and cloud shadow, as well as an other category for non-agricultural pixels.

\subsubsection{Model design}

\begin{figure*}[ht]
\vskip 0.2in
\begin{center}
\centerline{\includegraphics[width=14cm]{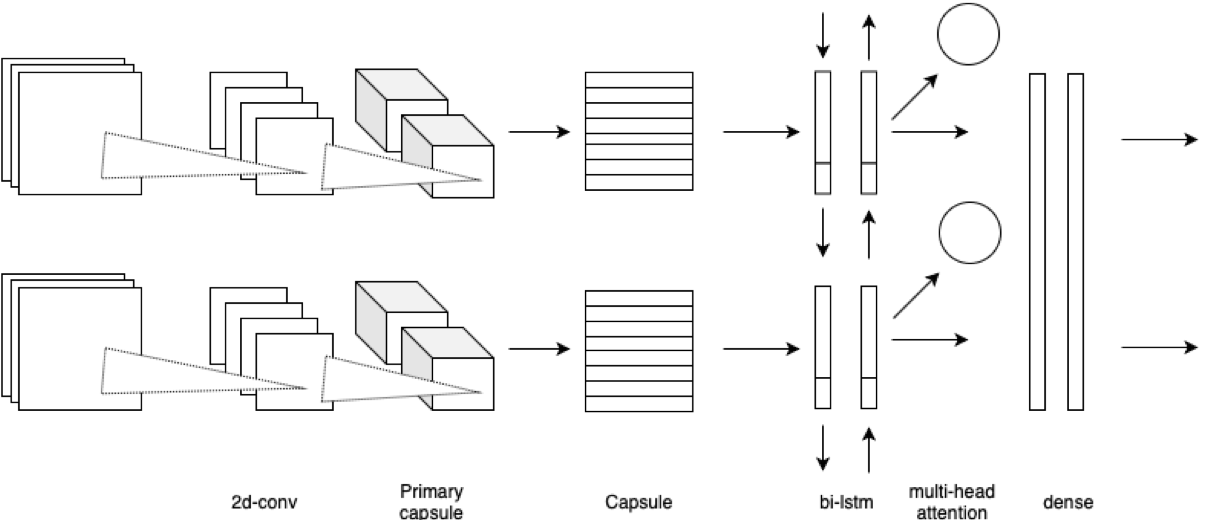}}
\caption{Diagrammatic overview of model architecture. Shape and activation specifics are outlined in Table \ref{model-table}.}
\label{model-figure}
\end{center}
\vskip -0.2in
\end{figure*}

The general model architecture (shown in Figure \ref{model-figure} and Table \ref{model-table}) begins with a 1x1 2D convolution over the 6 input channels with ReLU activation. The convolution output is passed to a primary capsule prior to the squashing activation function. The capsules are flattened and fed into a many-to-many bidirectional LSTM with 240 units, before being weighted with time-distributed attention and passed to two dense layers for classification.

\begin{table}[h!]
\vskip -0.1in
\caption{Architecture layer shape and activation.}
\label{model-table}
\vskip 0.15in
\begin{center}
\begin{small}
\begin{sc}
\begin{tabular}{lcccr}
\toprule
Layer & Shape & Activation\\
\midrule
Input &  26 x 3 x 3 x 6 & \\
2D-Conv  & 26 x 3 x 3 x 128 & ReLU \\
Primary Caps & 26 x 1280 x 10& ReLU \\
Capsule  & 26 x 23 x 10 & Squash\\
Bi-LSTM     &  26 x 480 & Tanh\\
Attention &26 x 480 & Tanh \\
Dense &26 x 512 & ReLu \\
Dense &26 x 23 & Softmax \\
\bottomrule
\end{tabular}
\end{sc}
\end{small}
\end{center}
\vskip -0.21in
\end{table}

This paper employs the capsule architecture set forth by Sabour, Frosst, and Hinton \yrcite{capsnet}. Instead of using margin and reconstruction loss and masking capsule layers with ground truth information, we use cross entropy loss for three reasons. First, unmasked capsule outputs provide less sparse input to the LSTM. Secondly, backpropagating margin loss through the LSTM to the capsule layer may reduce the amount of information passed between time steps in the LSTM, as the network will be incentivized to maintain more information about the individual time step, partially defeating the LSTM's purpose of creating a time-dependent representation. Finally, the use of capsule layers with cross entropy loss allows the architecture to create a flattened representation of the image that mitigates translation invariance issues in a similar manner as max pooling, without discarding any of the valuable and limited training information per sample. 

The output state of each time step from the LSTM layer is passed to distributed attention layers using the equations in (1-3) as introduced by Yang et al. \yrcite{inproceedings},

\begin{align}
u_{it} &= \text{tanh}(W_w h_{it}+b_s), \\
\alpha_i &= \frac{\text{exp}(u_{it}^Tu_w)}{\sum_t \text{exp}(u_{it}^T u_w}, \\
s_i &= \sum_t \alpha_{it}h_{it}
\end{align}

where $W$, $h$, and $b$ refer to the weight matrix, hidden state, and bias, respectively, and $\alpha$ is a learned parameter weighing each hidden state.  Three weight vectors are learned at each time step to weight and combine every time step into a vector that is then passed to two fully connected layers for classification.

Training was implemented on an Ubuntu 16.1 server running Tensorflow 1.5.1 with an implementation of capsule layers written in Keras 2.0.8. Training took approximately 12 hours on a Nvidia K80 GPU. A batch size of 50 was used with a learning rate of 0.01 and the Adam optimizer. Regularization methods, including dropout, batch normalization, and kernel regularization were tested. However, the model did not overfit and regularization did not improve generalization ability. Deeper networks, with stacked LSTM and/or 2D convolution layers, were also tested with a variety of skip connections to help gradient flow, but did not improve over the presented architecture.

\section{Results}

\begin{table}[h]
\caption{Model metrics comparing the proposed architecture with previous approaches. LSTM, RNN, CNN, and SVM results taken from Rubworm and Korner \yrcite{RuBwurm2017TemporalVM}. Metrics are weighted to account for class inbalance using the methodology in Rubworm and Korner \yrcite{RuBwurm2017TemporalVM}.}
\label{sample-table}
\vskip 0.15in
\begin{center}
\begin{small}
\begin{sc}
\begin{tabular}{lcccr}
\toprule
Model & Acc. & Precision & Recall & F-score \\
\midrule
\textbf{CapsAttn} & \textbf{88.0} & \textbf{87.5} & \textbf{85.6} & \textbf{86.5} \\
CapsLSTM & 86.7 & 85.1 & 83.8 & 84.5 \\
CNNAttn & 86.3 & 83.8 & 82.6 & 84.1 \\
CNN-LSTM & 85.6 & 83.9 & 82.1 & 82.3 \\
LSTM  & 74.3 & 78.4 & 74.5 & 75.3 \\
RNN    & 72.9 & 77.3 & 73.0 & 74.0 \\
CNN    & 64.3 & 59.2 & 57.2 & 56.7       \\
SVM     & 31.1 & 31.4 & 31.1 & 31.1 \\
\bottomrule
\end{tabular}
\end{sc}
\end{small}
\end{center}
\vskip -0.1in
\end{table}

Utilizing capsule layers and time distributed attention, the present architecture (hereafter referred to as CapsAttn) achieves state of the art results on crop type classification, improving upon the results reported by Rubworm and Korner \yrcite{RuBwurm2017TemporalVM} by a relative margin of over 80\%. Further metrics are provided in Table \ref{sample-table}, indicating that CapsAttn outperforms previous approaches. These results suggest that both capsule layers and attention can increase performance in low-resolution satellite classification. The CNN-LSTM network, with the output of a 1x1 2D convolution and 2D max pooling fed as input to an LSTM, also performed better than the LSTM in Rubworm and Korner \yrcite{RuBwurm2017TemporalVM}. This suggests that, while temporal information is important when classifying vegetation with Sentinel data, spatial relationships should not be discarded. The ablation study in Table \ref{sample-table} shows that capsule layers and distributed attention independently increase classification performance, but the combination of both performs the best. 

\begin{figure}[h]
\vskip 0.2in
\begin{center}
\centerline{\includegraphics[width=\columnwidth]{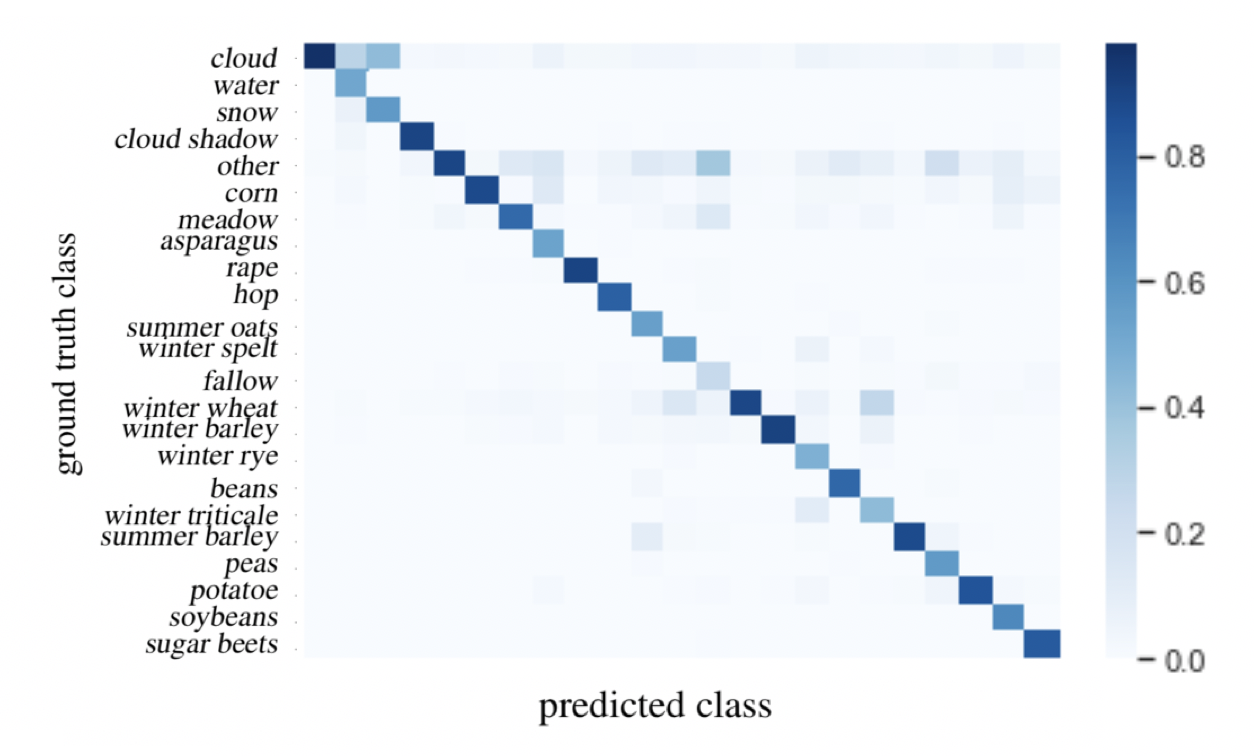}}
\caption{Confusion matrix weighted by percent accuracy of CapsAttn on the test data set.}
\label{icml-historical}
\end{center}
\vskip -0.2in
\end{figure}

Following the results of Rubworm and Korner \yrcite{RuBwurm2017TemporalVM} which demonstrated that time series classification greatly outperforms single-image classification for low resolution vegetation analysis, the present results suggest that distributed attention layers allow for more complicated time dependencies to be learned. Learning separate weights at each time step allows the architecture to differentiate between, for example, time dependencies between winter and early growing season, and spring and harvest season. Attention layers with shared weights did not see any significant improvement over the architecture without attention, indicating that learning individual weights at each time step is critical for the network to understand the complicated time dependencies present in agricultural growing patterns.

This paper diverges from previous remote sensing papers that have utilized capsule networks by eschewing margin and reconstruction loss. As outlined in the methodology, this was done due to the simplicity of the data, number of classes, and issues imposed by masking with a recurrent decoder. Other papers outside of remote sensing where time series or recurrent classification is necessary, such as in polyphonic sound detection \cite{liu_tang_song_dai_2018}, and text classification \cite{srivastava-etal-2018-identifying} have also found that unmasked capsule layers with cross entropy loss perform better than masked capsule layers with margin loss. As such, this paper suggests that the model architecture and data complexity should inform the choice of loss function for capsule layers in remote sensing. Overall, this paper introduces a state-of-the-art classification architecture for agricultural crops that allows for accurate classification with low resolution imagery, reducing error rate versus the previous state-of-the-art by nearly 50\%.


\nocite{langley00}

\bibliography{example_paper}
\bibliographystyle{icml2019}


\end{document}